\documentclass{article}

\usepackage[preprint]{tmlr}

\usepackage{microtype}
\usepackage{graphicx}
\usepackage{subcaption}
\usepackage{booktabs}
\usepackage{tabularx}
\usepackage{multirow}

\usepackage{hyperref}

\usepackage{amsmath}
\usepackage{amssymb}
\usepackage{mathtools}
\usepackage{amsthm}

\usepackage[capitalize,noabbrev]{cleveref}

\theoremstyle{plain}

\theoremstyle{definition}

\theoremstyle{remark}

\usepackage[textsize=tiny]{todonotes}

\title{How Proper Scoring Rules Shape LLM Forecasting}

\author{
\name Benjamin Turtel, Paul Wilczewski, and Kris Skotheim \\
\addr Lightning Rod Labs, \href{mailto:ben@lightningrod.ai}{ben@lightningrod.ai}
\AND
\name Ville A. Satop{\"a}{\"a} \\
\addr Technology and Operations Management, INSEAD, Fontainebleau, France, \href{mailto:ville.satopaa@insead.edu}{ville.satopaa@insead.edu}
\AND
\name Philip E. Tetlock \\
\addr Wharton School and School of Arts \& Sciences, University of Pennsylvania
}

\begin{document}

\maketitle

\begin{abstract}
This paper evaluates how reward function choice shapes the performance and behavior of LLM forecasters. We compare five proper scoring rules as training objectives for binary forecasts of resolved real-world events. Although the rules share the same theoretical incentive for truthful
probability reporting, the resulting models differ in calibration, probability
use, and estimated profiles of bias, information, and noise, with smaller
differences in aggregate accuracy and discrimination. The Brier-trained model has the lowest observed Brier score and highest AUC-ROC, while the log-trained model has the highest observed log score and lowest calibration error. Models with similar aggregate performance also reach that performance through different combinations of bias, information, and noise. Proper scoring rules therefore need not behave interchangeably as training
objectives. Reward choice may shape not only how well an LLM forecasts, but
how its forecasting errors are structured.
\end{abstract}

\section{Introduction}

Forecasting is a natural testbed for language-model reasoning. It requires models to combine incomplete and conflicting evidence, express uncertainty as a probability, and make predictions under limited information. It also permits objective evaluation. Once an event resolves, the forecast can be scored against the observed outcome.

This paper studies how reward choice relates to the behavior of LLM forecasters. We train models on binary forecasting questions derived from resolved news events. The questions span politics, geopolitics, economics, business, science, and sports with forecast horizons ranging from 7 to 90 days. Each model is optimized using Dr.~GRPO \citep{Liu25a}, a variant of Group Relative Policy Optimization (GRPO), with a proper scoring rule as the terminal reward. The model, data, optimizer, training steps, and rollout budget are held fixed across conditions.

We compare five reward functions: the logarithmic, Brier, spherical,
Beta$(2,8)$, and Beta$(8,8)$ scores. All five are strictly proper scoring
rules and therefore incentivize truthful probability reporting in expectation
\citep{gneiting2007strictly}. The two beta scores belong to the beta family of
proper scoring rules introduced by \citet{buja2005loss}. Although all
five scores share the same truthfulness property, they differ in curvature,
boundedness, tail behavior, and the probability regions they emphasize. Under
Dr.~GRPO, these differences induce different relative advantages among sampled
forecasts.

We evaluate the resulting models using aggregate forecasting score, calibration, discrimination, probability-scale use, and the bias--information--noise decomposition of \cite{satopaa2021bias}. Most reward-trained variants improve over the untrained GPT-OSS-120b model, but their performance profiles differ. The Brier-trained model achieves the best Brier score and AUC-ROC, while the log-trained model achieves the best log score and lowest expected calibration error. Models with similar aggregate performance also show different estimated contributions from bias, information, and noise. The BIN decomposition therefore reveals distinctions that are not apparent from aggregate forecasting metrics alone.

Our contributions are:
\begin{itemize}
\item We provide a controlled finite-training comparison of five proper scoring rules used as terminal rewards for LLM forecasting.

\item We show that reward choice is associated with differences in aggregate
performance, calibration, discrimination, and probability-scale use, while BIN
analysis reveals distinct estimated contributions from bias, information, and
noise.

\item We show that proper scoring rules with the same population-level incentive
are associated with different learned forecasters in our finite group-relative
policy optimization experiments.
\end{itemize}

\section{Related Literature}

\noindent\textbf{Language-model forecasting.}
Language models have been evaluated on temporally grounded forecasts of real-world events. Autocast introduced forecasting questions paired with time-indexed news evidence \citep{zou2022forecasting}. Later systems improved performance through retrieval, structured reasoning, and forecast aggregation \citep{halawi2024approaching}. ForecastBench evaluates models on unresolved questions to reduce temporal leakage and compares their forecasts with those of human forecasters \citep{karger2024forecastbench}. Our work is most closely related to Future-as-Label, which uses outcomes observed after an information cutoff as supervision for forecasts based only on pre-cutoff evidence \citep{turtel2026futureaslabel}. Whereas prior work largely focuses on forecasting systems, supervision, or evaluation, we hold the assigned training setup fixed and study the role of the reward function.

\noindent\textbf{Proper scoring rules and forecast evaluation.}
Proper scoring rules incentivize truthful probabilistic reports in expectation. However, they differ in curvature, boundedness, tail behavior, and the probability regions they emphasize \citep{gneiting2007strictly}. Binary proper losses can be characterized by weight functions over the probability interval, which motivates objectives that place different emphasis on different forecast regions \citep{buja2005loss,reid2010composite}. We study these differences under Group Relative Policy Optimization, where advantages are determined by comparing rewards across responses sampled for the same input \citep{shao2024deepseekmath,Liu25a}. More broadly, reinforcement-learning post-training has shown that the specification of the reward signal can materially shape language-model behavior, from preference alignment to reasoning performance \citep{ouyang2022training,guo2025deepseekr1}, while work on reward over-optimization shows that optimizing a reward proxy need not preserve the behavior ultimately of interest \citep{moskovitz2023confronting}. Our setting differs in that the rewards are not learned proxies: each is a strictly proper scoring rule with the same population-optimal report. Nevertheless, their different geometries can generate different finite-sample learning signals under group-relative optimization.

Forecast evaluation considers both aggregate accuracy and the structure of probabilistic errors. Classical Brier-score analysis decomposes performance into reliability, resolution, and uncertainty \citep{murphy1973new}, while the bias--information--noise decomposition characterizes differences in terms of systematic bias, valid information, and nonsystematic noise \citep{satopaa2021bias}. Calibration metrics and reliability diagrams are also widely used to evaluate probabilistic neural-network predictions \citep{guo2017calibration}.

\section{Method}

\subsection{Forecasting Task}

We consider binary forecasting questions of the following form:

\begin{quote}
    \emph{Will event $E$ occur by resolution time $s$?}
\end{quote}

Examples include:
\begin{quote}
\emph{Will the Trump administration officially announce a suspension or termination of the 50\% tariffs on imports from India before March 1, 2026?}

\emph{Will André Ventura win the Portuguese presidential runoff election held on February 8, 2026?}

\emph{Will the U.S. Consumer Price Index for All Urban Consumers (CPI-U) show a year-over-year increase of at least 4.0\% in the Bureau of Labor Statistics' March 2026 report?}
\end{quote}

Each question resolves to \emph{Yes} if the specified event occurs by the resolution time and to \emph{No} otherwise.

For each question, the model receives information available at prediction time $t < s$ and outputs a probability $p \in (0,1)$ that the event will occur. After the event resolves, the outcome is recorded as $y \in \{0,1\}$ where a value of one indicates the event occurred while a value of zero indicates the event did not occur.

During training, the model is presented with one forecasting question and the information that would have been available at prediction time $t$. It generates its reasoning and a final probability forecast. The forecast is then scored against the event's resolved outcome. The details of how multiple model responses are sampled and used for optimization are described below.

\subsection{Temporal Masking and Outcome Resolution}

We follow the future-as-label setup of \citet{turtel2026futureaslabel}. The predictor receives only evidence available at or before the cutoff $t$. The outcome is determined after resolution time $s > t$ using evidence unavailable to the predictor. 

For example, consider the question \emph{``Will the Trump administration officially announce a suspension or termination of the 50\% tariffs on imports from India before March 1, 2026?''} The forecasting cutoff for this question is January 15, 2026. The predictor has access to news articles published on or before January 15, 2026, while the resolved outcome is determined from news articles published after January 15, 2026.

The predictor is the language model being trained. It observes $x_t$, samples a reasoning trajectory $\tau$, and produces a final forecast $p_\theta(\tau, x_t)$. A fixed external resolver assigns the binary outcome $y_s$ using post-$t$ evidence. It does not observe the forecast, evaluate the reasoning trajectory, or provide preference judgments.

Training is performed offline on resolved events, but temporal masking preserves the information constraints in place at prediction time. Supervision comes from the resolved outcome rather than from judgments about the model's reasoning.

\subsection{Reward Functions and Evaluation Metrics}
\label{sec:reward-functions}

For each resolved training example, the model's final forecast receives a
terminal reward $S_m(p,y)$, where $m$ indexes the reward function,
$p \in (0,1)$ is the forecast, and $y \in \{0,1\}$ is the resolved outcome.
We compare five proper scoring rules: log, Brier, spherical,
Beta$(2,8)$, and Beta$(8,8)$. Rewards depend only on the final parsed
probability and the outcome of the event. The reward functions evaluated in our
experiments are defined as follows.

\noindent\textbf{Logarithmic reward.}
We define the logarithmic reward as the logarithmic score \citep{Good1952}:
\begin{equation}
S_{\mathrm{log}}(p,y)
=
y \log p
+
(1-y)\log(1-p).
\end{equation}

\noindent\textbf{Brier reward.}
Because rewards are maximized during training, we use the negative of the
standard Brier score \citep{Brier1950}. Specifically,
\begin{equation}
S_{\mathrm{Brier}}(p,y)
=
-(p-y)^2.
\end{equation}
Thus, maximizing the Brier reward is equivalent to minimizing the conventional
Brier score. For evaluation, we report the conventional Brier score, $(p-y)^2$, so lower
values indicate better performance; this is the negative of the Brier reward
used during training.

\noindent\textbf{Spherical reward.}
We define the spherical reward as the spherical score \citep{gneiting2007strictly}
\begin{equation}
S_{\mathrm{spherical}}(p,y)
=
\frac{
yp+(1-y)(1-p)
}{
\sqrt{p^2+(1-p)^2}
}.
\end{equation}

\noindent\textbf{Beta-family rewards.}
We define the beta-family rewards as the beta-family of proper scoring rules \citep{buja2005loss}:
\begin{equation}
S_{\alpha,\beta}(p,y)
=
-y
\int_{p}^{1}
(1-q)w_{\alpha,\beta}(q)\,dq
-
(1-y)
\int_{0}^{p}
q\,w_{\alpha,\beta}(q)\,dq,
\end{equation}
where the normalized weight function is
\begin{equation}
w_{\alpha,\beta}(q)
=
\frac{
q^{\alpha-1}(1-q)^{\beta-1}
}{
B(\alpha,\beta)
},
\end{equation}
and $B(\alpha,\beta)$ denotes the beta function. The parameters $\alpha$ and
$\beta$ determine where the scoring rule places greater weight across the
probability range. We evaluate Beta$(2,8)$ and Beta$(8,8)$: Beta$(2,8)$ places
greater weight on errors in lower-probability regions, whereas Beta$(8,8)$
concentrates its weight near $0.5$. Beta$(1,1)$ corresponds to a uniform weight
function and recovers the Brier reward up to a positive multiplicative constant.
For numerical stability, beta-family forecasts are clipped to
$[10^{-3},1-10^{-3}]$ before reward computation.

\begin{figure}[t!] \centering \includegraphics[width=.7\linewidth]{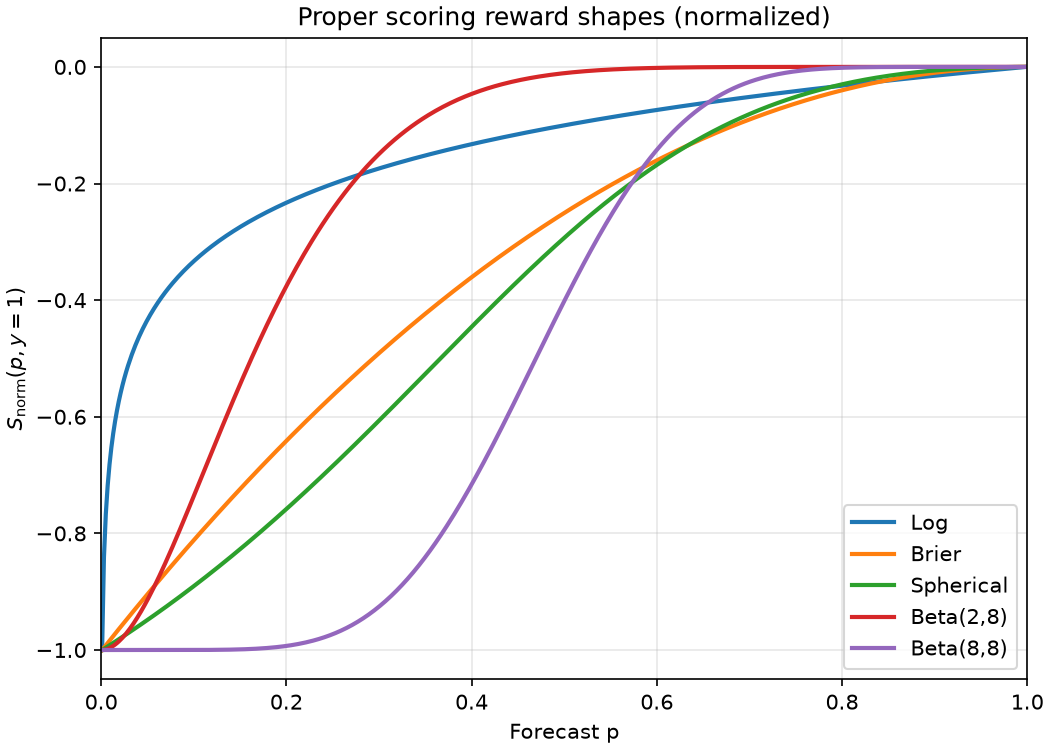} \caption{Normalized reward vs. forecast probability for five proper scoring rules ($y=1$).} \label{fig:reward-function-shape} \end{figure}

Figure~\ref{fig:reward-function-shape} was generated by evaluating each reward
on a dense grid of forecast probabilities for a positive outcome ($y=1$) and
normalizing each reward independently to the interval $[-1,0]$. Although all five rewards are proper, they differ in geometry and in the
probability regions they emphasize. The log
and Beta$(2,8)$ rewards are more sensitive to low-probability forecasts, whereas
Beta$(8,8)$ concentrates its variation near $p=0.5$; the Brier and spherical
rewards vary more smoothly across the range. These differences can produce
distinct learning signals under finite policy optimization.

Beyond these scoring rules, we report complementary measures of calibration,
discrimination, and forecast structure.

\noindent\textbf{Expected calibration error.}
Expected calibration error measures the difference between predicted
probabilities and observed outcome frequencies. We divide forecasts into
$B=10$ bins and compute
\begin{equation}
\operatorname{ECE}
=
\sum_{b=1}^{B}
\frac{\lvert I_b\rvert}{N}
\left\lvert
\frac{1}{\lvert I_b\rvert}
\sum_{i \in I_b} y_i
-
\frac{1}{\lvert I_b\rvert}
\sum_{i \in I_b} p_i
\right\rvert,
\end{equation}
where $I_b$ is the set of forecasts in bin $b$. Lower values indicate better
calibration.

\noindent\textbf{AUC-ROC.}
AUC-ROC measures how well the forecasts rank positive outcomes above negative
outcomes. It depends on the ordering of predictions rather than their absolute
calibration. Higher values indicate better discrimination.

\noindent\textbf{Bias--information--noise decomposition.}
We use the bias--information--noise decomposition of
\citet{satopaa2021bias} to characterize differences in expected Brier score.
The decomposition separates systematic bias, valid information, and
non-systematic noise. It therefore distinguishes changes in average probability
levels from changes in outcome-relevant information and forecast variability.
Because the decomposition is model-based, its components describe an estimated
rather than observed difference in Brier score.

\subsection{Training details}

We optimize the forecasting policy using the Dr.~GRPO formulation of
\citet{Liu25a}. For each forecasting state $x_t$, the model samples a group of
$K$ reasoning trajectories $\{\tau_i\}_{i=1}^{K}$. Each trajectory terminates
in a parsed probability forecast $p_i$ and receives reward
$S_m(p_i,y)$ according to scoring rule $m$ and realized outcome $y$.

Dr.~GRPO constructs a relative advantage by comparing each trajectory's reward
with the rewards of other trajectories sampled from the same forecasting
state. In our setting, the advantage for trajectory $i$ is

\begin{equation}
A_i
=
S_m(p_i,y)
-
\frac{1}{K}
\sum_{j=1}^{K}
S_m(p_j,y).
\end{equation}

Thus, a trajectory receives positive advantage when its forecast scores better
than the average forecast in its rollout group and negative advantage when it
scores worse. The policy update increases the likelihood of trajectories with
positive advantage and decreases the likelihood of trajectories with negative
advantage. The group mean therefore acts as a state-specific baseline: it
provides a relative learning signal while leaving the underlying reward
defined by the scoring rule.

Following Dr.\ GRPO, we do not divide the centered rewards by the within-group reward standard deviation, and we apply no KL penalty. Dr.\ GRPO introduces these modifications to address biases that can arise in standard GRPO; in our setting, they also serve an additional forecasting-specific purpose. We want the optimization objective to preserve the incentives induced by a proper scoring rule. Under a proper scoring rule, a forecaster maximizes expected reward by reporting its true predictive probability. Dividing rewards by a group-dependent standard deviation makes the scale of the learning signal depend on the particular forecasts and rewards sampled within a rollout group, rather than preserving the scoring rule on its original scale. Likewise, adding a KL penalty changes the objective from maximizing expected scoring-rule reward to trading off forecast quality against proximity to a reference policy. We therefore retain only group centering, which provides a state-specific baseline while keeping the relative learning signal directly tied to the proper scoring-rule reward. The proper scoring rule solely determines the reward, and we assign no additional process-level reward to the reasoning trajectory.

Invalid outputs receive a fixed penalty on the scale of the corresponding
scoring rule: $0$ for the spherical reward, $-1$ for the Brier and beta-family
rewards, and $-8$ for the log reward. These values are chosen so that an invalid
output receives a reward at least as poor as the worst valid forecast under the
corresponding reward function, while remaining on a comparable reward scale.
These penalties are matched to each rule's own reward scale, not to the
group-relative advantages that Dr.~GRPO propagates, so an invalid output
represents a larger deviation from the group mean under some rules than others.
Because invalid outputs are rare
(Table~\ref{tab:model-results-5pooled}), we do not expect this to materially
affect the comparison. For numerical stability all probabilities are clipped to
$[0.001,0.999]$. The floor bounds the log score at approximately $-6.9$, keeping
the worst valid forecast above the invalid-output penalty but penalizing
low-tail overconfidence less than the unclipped rule would.

We train GPT-OSS-120B with Dr.~GRPO using rank-32 LoRA adapters under five reward functions, one for each reward defined in Section~\ref{sec:reward-functions}. Training is implemented with Tinker \citep{tinker}, which provides LoRA-based infrastructure for reinforcement-learning post-training. We compare each trained model with the unmodified GPT-OSS-120B model prior to training, which we refer to as the base model. Across all training runs, we hold fixed the model initialization, training data, optimizer, number of training steps, rollout budget, and hyperparameters, varying only the reward function. Hyperparameters were selected within the ranges recommended in the Tinker documentation and held fixed across conditions to isolate the effect of reward choice. Table~\ref{tab:training-hyperparameters} reports the shared training configuration. The primary five-reward comparison uses a single training seed per condition. We additionally repeat the Log and Brier conditions across three seeds as a targeted robustness analysis in Appendix~\ref{app:seed-robustness}.

\begin{table}[htbp]
\centering
\caption{Shared hyperparameters across reward-function variants.}
\label{tab:training-hyperparameters}
\begin{tabular}{lc}
\toprule
Hyperparameter & Value \\
\midrule
Batch size & 32 \\
Maximum response length & 16{,}384 \\
Training steps & 200 \\
Adapter & LoRA \\
Rank & 32 \\
Rollouts per question & 8 \\
Learning rate & $2 \times 10^{-5}$ \\
Warmup ratio & 0.1 \\
\bottomrule
\end{tabular}
\end{table}

\subsection{Data}

We follow the question-construction, temporal-masking, outcome-resolution, and
leakage-prevention procedures of \citet{turtel2026futureaslabel}. Questions are constructed from information available as of a prediction cutoff
and concern events that resolve only after that cutoff. Examples that cannot be
resolved unambiguously are excluded. 

Using this procedure, we construct an expanded dataset of binary forecasting
questions from news events between July 2024 and January 2026 using Lightning
Rod's proprietary SDK. The dataset contains 8{,}041 examples spanning politics,
geopolitics, economics, business, science, and sports, with 7{,}076 used for
training and 965 reserved for held-out evaluation. The positive outcome rate is
27.0\% in the training set and 28.5\% in the evaluation set.

Each example contains a temporally masked question context and a resolved binary outcome; representative examples are provided in Appendix~\ref{app:forecast_examples}. All reward variants use the same training and evaluation splits, and the held-out set is used only for final evaluation. All models are evaluated using the same prompt, decoding configuration, and probability-extraction procedure across conditions. For each held-out question, we sample five independent responses at temperature $1.0$. Our primary evaluation pools performance across all five sampled forecasts. Brier score, ECE, and log score are computed over all sampled forecast--outcome
pairs, while AUC-ROC is computed separately for each rollout and averaged across
the five rollout sets. The BIN decomposition is fit jointly to all 4{,}825 pooled
forecast--outcome observations. Median-of-five and single-sample evaluations are reported as robustness analyses in Appendix~\ref{app:detailed-results}.

\section{Results}
\subsection{Aggregate Performance}

Table~\ref{tab:model-results-5pooled} reports performance on the 965 held-out questions under the primary pooled five-rollout evaluation. Higher values are better for log score and AUC-ROC, while lower values are better for Brier score, ECE, and the rate of missing (unparseable) forecasts. All reward-trained variants have lower observed Brier score and ECE than the base model; relative to base, the log, Brier, spherical, and Beta$(2,8)$ variants improve Brier score at the 1\% significance level, and Beta$(8,8)$ at the 5\% level. The Brier-trained model achieves the best Brier score and AUC-ROC, while the log-trained model achieves the best log score and lowest ECE. Missing forecasts are rare throughout ($\leq 0.27\%$). 

\begin{table}[t!]
\centering
\caption{Held-out forecasting performance by reward variant.}
\label{tab:model-results-5pooled}
\begin{tabular}{lrrrrrrr}
\toprule
Reward variant & Log score $\uparrow$ & Brier $\downarrow$ & Sig. & ECE $\downarrow$ & AUC-ROC $\uparrow$ & Avg tokens & Missing (\%) \\
\midrule
Log & \textbf{-0.5132} & 0.1653 & *** & \textbf{0.0434} & 0.7407 & 454.3 & 0.00 \\
Brier & -0.5185 & \textbf{0.1648} & *** & 0.0568 & \textbf{0.7511} & 1089.4 & 0.00 \\
Spherical & -0.5273 & 0.1680 & *** & 0.0538 & 0.7353 & \textbf{362.8} & 0.02 \\
Beta$(2,8)$ & -0.5223 & 0.1677 & *** & 0.0449 & 0.7372 & 790.4 & 0.00 \\
Beta$(8,8)$ & -0.5536 & 0.1731 & ** & 0.0954 & 0.7366 & 2691.9 & 0.27 \\
Base & -0.5585 & 0.1861 & & 0.1099 & 0.7248 & 662.6 & 0.02 \\
\bottomrule
\end{tabular}
\end{table}

We quantify uncertainty using paired, question-level bootstrap confidence intervals with 2{,}000 resamples, retaining all five sampled forecasts for each resampled question. Arrows in Table~\ref{tab:model-results-5pooled} indicate the preferred direction of each metric, and bold text indicates the best-performing value in each column. The significance column (Sig.) reports paired comparisons of Brier score against the base model, with $^{***}$, $^{**}$, and $^{*}$ denoting significance at the 1\%, 5\%, and 10\% levels, respectively.

The reward conditions also differ in the number of tokens generated during training. The longest-output condition generates approximately seven times as many tokens as the shortest-output condition despite identical step and rollout budgets. The conditions therefore differ in realized token-level compute. We discuss this limitation in Section~\ref{sec:limitations}.

\begin{figure}[t!] \centering \includegraphics[width=.8\linewidth]{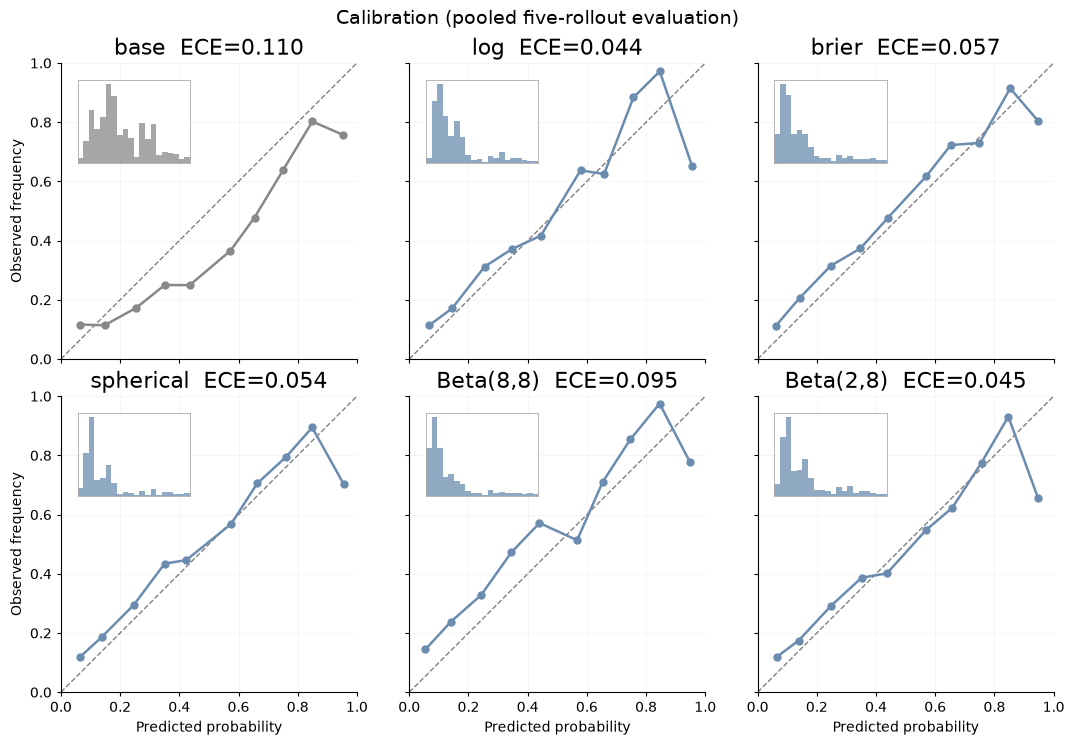} \caption{Held-out calibration curves and forecast distributions by reward variant.} \label{fig:calibration-plots} \end{figure}

Figure~\ref{fig:calibration-plots} compares calibration across the held-out forecasts. Each calibration curve groups forecasts into equally spaced probability bins and plots the observed outcome frequency within each bin against predicted probability. The dashed diagonal indicates perfect calibration, while the inset histograms show the corresponding distribution of forecast probabilities. Relative to the base model, the reward-trained variants generally track the diagonal more closely. Their forecast distributions also place substantially more mass at lower predicted probabilities than the base model.

\subsection{Bias--Information--Noise Decomposition}

We use the bias--information--noise (BIN) decomposition to characterize the
sources of Brier-score differences relative to the base model. The
decomposition separates contributions from systematic bias, outcome-relevant
information, and nonsystematic forecast noise. Because BIN is estimated under a
latent-normal model of the forecasts, the components sum to the model-implied
reduction in mean Brier score relative to the untrained base model rather than
to the observed difference reported in
Table~\ref{tab:model-results-5pooled}; the two differ by roughly one to two
percentage points of base Brier in our results. We report each component as a
percentage of the model-implied base Brier score. Positive values contribute to
improvement, whereas negative values offset it.

\begin{figure}[htbp]
\centering
\includegraphics[width=.8\linewidth]{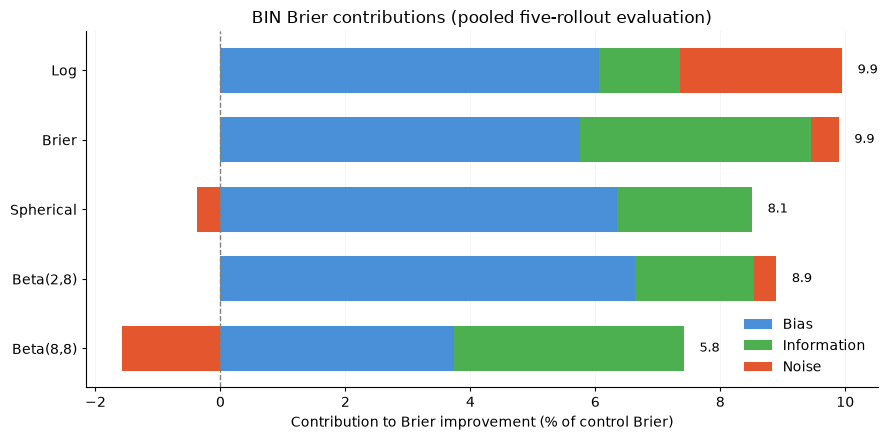}
\caption{BIN contributions relative to the base model under pooled five-rollout evaluation.}
\label{fig:bin-analysis}
\end{figure}

Figure~\ref{fig:bin-analysis} shows that the reward variants improve through
different combinations of bias, information, and noise under the primary pooled
five-rollout evaluation. Beta$(2,8)$ has the largest estimated bias contribution,
while the Brier variant has the largest estimated information contribution and
little contribution from noise. The log variant shows a comparatively large
positive noise contribution, whereas the spherical and Beta$(8,8)$ variants
show negative noise contributions that offset gains from bias and information.
These differences illustrate that similar aggregate Brier-score improvements
can arise from distinct forecasting profiles. The Brier information advantage
and log noise contribution are stable across evaluation methods, while the beta
variants are more sensitive to the choice of evaluation method. Detailed
numeric results and corresponding decompositions under single-sample and
median-of-five evaluation are reported in Appendix~\ref{app:detailed-results}.

\paragraph{BIN pairwise contrasts.}
For the primary pooled five-rollout evaluation, each contrast compares two groups
(control, treatment) using Hamiltonian Monte Carlo (2000 warmup and 4000
post-warmup iterations; seed 1). Reported quantities are posterior probabilities
formed within each MCMC draw and averaged across draws. For groups $A$
(treatment) and $B$ (control), draw-wise contrasts are
\[
D_{\mathrm{bias}}^{(s)} = |\mu_A^{(s)}| - |\mu_B^{(s)}|, \quad
D_{\mathrm{info}}^{(s)} = \gamma_A^{(s)} - \gamma_B^{(s)}, \quad
D_{\mathrm{noise}}^{(s)} = \delta_A^{(s)} - \delta_B^{(s)}.
\]
Bias uses absolute magnitude because $\mu$ is signed (over- vs.\
under-estimation); information $\gamma\in[0,1]$ and noise $\delta>0$ are
already one-sided. We report $P(|\mu_A|<|\mu_B|)$,
$P(\gamma_A>\gamma_B)$, and $P(\delta_A<\delta_B)$.

The pairwise contrasts reinforce the differences in forecasting profiles across
reward variants. For bias, the log and Beta$(2,8)$ variants have the strongest
posterior support, with little evidence separating the two. For information,
Brier and Beta$(8,8)$ dominate the other variants, while their pairwise
difference is comparatively uncertain. The log variant has the strongest noise
profile, with near-unit posterior probability of lower noise than every other
reward-trained variant. In contrast, Beta$(8,8)$ has the weakest noise profile.

\begin{figure}[htbp]
\centering
\includegraphics[width=\linewidth]{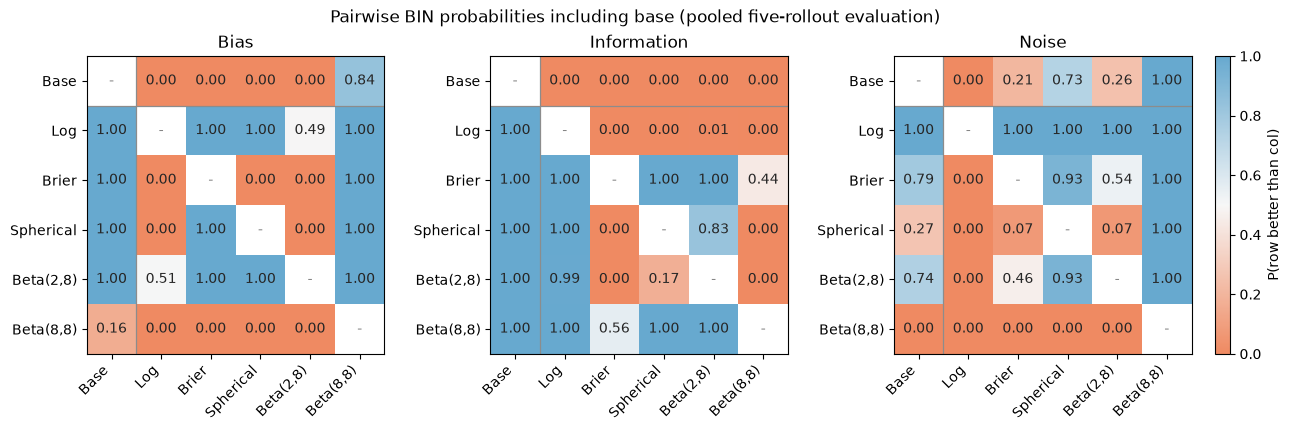}
\caption{Pairwise BIN posterior probabilities under pooled five-rollout evaluation.}
\label{fig:bin-significance}
\end{figure}

\section{Discussion and Limitations}
\label{sec:limitations}

This study compares how five proper scoring rules behave as terminal rewards
for LLM forecasting under a shared finite-training setup. The reward-trained
models generally outperform the base model, but their performance profiles
differ. Variants with similar Brier scores show clearer differences in
calibration, discrimination, probability-scale use, and the estimated balance
of bias, information, and noise. Aggregate scores therefore capture only part
of the variation among the learned forecasters.

These results illustrate a practical distinction among theoretically proper
objectives. Each scoring rule incentivizes truthful probability reporting in
expectation, but the rules generate different reward distributions and
group-relative advantages during finite policy optimization. In our
experiments, these differences are associated with distinct learned
forecasting profiles. This does not challenge the propriety of the scoring
rules. Rather, it shows that their empirical behavior as LLM training
objectives can differ under limited data, sampled rollouts, and a fixed
optimization procedure.

The composition of forecast error can matter even when aggregate performance is
similar. A forecaster with low systematic bias but greater nonsystematic noise
may be attractive when repeated forecasts can be aggregated and idiosyncratic
errors reduced. A lower-noise forecaster may instead be preferable when
individual forecasts are acted on directly and consistency is important.
Information gains may be especially valuable when forecasts are used for
ranking or prioritization. In this setting, reward choice appears to provide a
potential lever over these trade-offs. Determining whether particular scoring
rules reliably induce particular BIN profiles will require experiments across
models, datasets, and optimization settings.

Differences in BIN profiles also suggest a role for reward diversity in
forecast ensembles. An ensemble need not consist only of copies of the
individually best-performing forecaster. Models trained with different rewards
may contribute complementary error profiles, even when their standalone scores
are similar. Future work could test whether deliberately combining such models
improves ensemble performance, and whether an aggregation policy can itself be
learned from forecasting feedback. A related direction is joint optimization
of ensemble members and their aggregator under a proper scoring objective.

The analysis has several limitations. We study one base model, one dataset of
binary news-event forecasts, and one shared hyperparameter configuration. The primary five-reward comparison uses a single training seed per condition, so some observed differences may reflect training stochasticity. A targeted three-seed replication of the Log and Brier conditions is reported in Appendix~\ref{app:seed-robustness}; while small aggregate-performance differences are seed-sensitive, the qualitative BIN distinction between the two reward conditions is preserved. Hyperparameters are held fixed
across reward functions rather than tuned separately for each objective. The
results therefore compare rewards under a common optimization setup, not each
reward at its individually optimized configuration. Generalization may also
depend on the forecasting domain, event base rate, optimization procedure, or
training budget. The asymmetric Beta$(2,8)$ reward may be particularly
sensitive to the outcome distribution. Reward conditions also differ in
realized token generation despite sharing the same step and rollout budgets,
so they do not use identical token-level compute.

The results should therefore be interpreted as a controlled analysis of
reward-function behavior in this setting, not as evidence that one scoring rule
is universally preferred. Future work should test whether the observed relationships generalize across models, datasets, base rates, and training seeds for all reward conditions. It would also be useful to compare these learned BIN profiles with those
of human forecasters. Humans may respond differently, or less strongly, to the
choice of scoring rule, raising the question of what forecasting objective
their behavior implicitly reflects. Understanding these differences could help identify reward functions that better align LLM forecasting behavior with, or complement, human forecasters.

\bibliographystyle{tmlr}
\bibliography{forecasting-behavior}

\newpage

\appendix

\section{Example Forecasting Inputs}
\label{app:forecast_examples}

This appendix provides two examples of the inputs presented to the model. Each example contains the forecasting question, prediction date, resolution criteria, close date, and temporally available news context. The same forecasting instructions and output format were used for all questions and are shown once below.

\subsection{Forecasting Instructions}

\begin{quote}
\small
You are a forecasting analyst. Given a question about a future event, your task is to produce a well-calibrated forecast for how it will resolve by the close date.

Treat established facts in the context as true and reason forward from them. Treat predictions, projections, or expert opinions within the context as evidence to weigh, not ground truth. The question date may fall after your knowledge cutoff; rely on the context for the state of the world at that time.

If context is unavailable or insufficient, do not invent facts. Use only durable background knowledge, acknowledge the limitation, and widen uncertainty.

\medskip
\noindent\textbf{Answer format:}
Think carefully in English about your answer and output your final prediction (a float between 0.0 and 1.0) between \texttt{<answer></answer>} tags. Example outputs are provided only to illustrate the format and should not be considered baselines, e.g., \texttt{<answer>0.75</answer>}.
\end{quote}

\subsection{Example 1: Russia--Ukraine Ceasefire}

\noindent\textbf{Question.}
Will a formal ceasefire agreement between Russia and Ukraine be signed by representatives of both nations before March 1, 2026?

\medskip
\noindent\textbf{Prediction date.}
December 3, 2025.

\medskip
\noindent\textbf{Resolution criteria.}
The question resolves to \emph{Yes} if a written ceasefire agreement intended to halt hostilities in the ongoing conflict is signed by authorized representatives of the Russian and Ukrainian governments on or before February 28, 2026. Informal ``understandings'' or unilateral declarations of a pause in fighting do not qualify; the agreement must be a bilateral or multilateral document confirmed by official state sources or major international news outlets.

\medskip
\noindent\textbf{Close date.}
March 1, 2026.

\medskip
\noindent\textbf{Context.}
Recent news articles relevant to the question were summarized as follows:

\begin{enumerate}
\small

\item \textbf{Trump pushes Ukraine, Russia on peace deal, but key issues are unresolved} 
(\emph{Washington Post}, November 25, 2025; relevance: 5.0).

Trump said talks with Ukraine and Russia were making progress and that only ``a few remaining points of disagreement'' stood after a week of rapid negotiations. He projected optimism but said he would meet leaders of the warring countries only once they were in the final phases of a deal. Despite the closer U.S.--Ukraine cooperation after long Geneva discussions, the two sides still diverge on key terms---especially whether Kyiv should surrender additional territory to Moscow. Overnight Russian strikes killed people in Kyiv on November 25.

\item \textbf{Russian Offensive Campaign Assessment, November 30, 2025}
(\emph{Institute for the Study of War}, November 30, 2025; relevance: 4.0).

On November 30, 2025, U.S. Secretary of State Marco Rubio said the U.S.-proposed peace plan seeks to ensure Ukraine's independence, sovereignty, and long-term prosperity. Ukrainian NSDC Secretary Rustem Umerov reported ``substantial progress'' with the U.S. and that negotiations continue on a joint peace framework; Kremlin spokesperson Dmitry Peskov said Putin would host U.S. envoy Steve Witkoff in Moscow before Putin's India trip. Meanwhile, Russian milbloggers and ultranationalist figures argued the Kremlin will likely reject any ceasefire or U.S. plan as it conflicts with Russia's war aims. ISW assesses Russian battlefield ``victory'' claims as exaggerated.

\item \textbf{Ukraine peace talks shift to Moscow---the big unknown is whether Putin will play ball}
(\emph{CNBC}, December 1, 2025; relevance: 4.0).

U.S. special envoy Steve Witkoff is traveling to Moscow to meet Vladimir Putin to discuss a U.S.-backed 19-point Ukraine peace plan. Ukraine has initially backed the proposals, but no final agreement has been reached, putting the next step in Russia's hands. Putin has cautiously said the plan could form the basis for future agreements, while also praising Russian advances and suggesting fighting would stop only if Ukrainian forces withdraw from key areas. Russian information sources appear skeptical Putin will compromise on territorial demands. U.S. officials call the process ``delicate.''

\item \textbf{How are Negotiations for a Ceasefire Between Russia and Ukraine Going?}
(\emph{NPR}, November 24, 2025; relevance: 4.0).

U.S. and Ukrainian officials say they are making progress toward ending the Russia--Ukraine war, but European allies feel excluded from the U.S.-backed plan. They argue it risks appeasing Russia rather than addressing broader concerns. The episode includes reaction from Ukrainians about the state of the negotiations and from a German diplomat on what the plan is missing. The discussion centers on whether talks can produce a ceasefire agreement and what key elements remain unresolved among participating and affected parties.

\item \textbf{Ukraine agrees to peace proposal, with only ``minor details'' to settle, official says, but no word from Russia}
(\emph{CBS News}, November 26, 2025; relevance: 4.0).

A U.S. official says Ukraine has agreed to a Russia--Ukraine peace deal brokered by the Trump administration, with only ``minor details'' remaining. Ukraine's national security adviser Rustem Umerov says negotiators reached ``core terms'' amid ongoing talks in Abu Dhabi involving U.S., Ukrainian and Russian officials. Trump says only ``a few remaining points'' differ and directed Steve Witkoff to meet Putin in Moscow. Russia has not commented on any agreement's details; Foreign Minister Sergey Lavrov calls it ``premature'' and says Russia expects an updated proposal reflecting Alaska understandings.

\item \textbf{U.S. pushing Ukraine to sign peace deal by Thanksgiving or lose support}
(\emph{Washington Post}, November 21, 2025; relevance: 4.0).

The White House is pressuring Ukraine to sign its new peace proposal by Thanksgiving or lose U.S. support in its war with Russia, according to five people familiar with the talks. U.S. officials are sending ``signals'' that if Kyiv does not quickly sign the proposal---drafted by special envoy Steve Witkoff---then ``everything could be off the table.''

\item \textbf{Putin sees U.S. peace plan as a starting point as he warns Ukraine's army to withdraw}
(\emph{PBS}, November 27, 2025; relevance: 3.0).

Putin said U.S. proposals for ending the war could be a basis for talks and urged Ukrainian forces to pull back or be overrun; he said hostilities would cease only if Ukraine withdraws from territories it occupies, otherwise Russia would achieve it by force. Putin called Trump's plan ``issues for discussion,'' not a draft agreement. Russia has demanded full withdrawal from Donetsk, Luhansk, Kherson, and Zaporizhia, including areas Russia does not occupy, and also wants to prevent NATO entry and Western troops. U.S. envoy Steve Witkoff is set to visit Moscow; Dan Driscoll may go to Kyiv. Fighting continues alongside these efforts.

\item \textbf{Ukraine updates: US-Russia talks reported in Abu Dhabi}
(\emph{DW}, November 25, 2025; relevance: 3.0).

Ukrainian strikes hit Russia's Krasnodar border region and the port city of Taganrog, while Russian drones and missiles bombarded Kyiv, killing at least one and injuring others and damaging residential and energy infrastructure. Diplomatically, U.S.--Russia and U.S.--Ukraine negotiations tied to Trump's 28-point peace plan continued, with Steve Witkoff reported to meet Vladimir Putin the following week in Moscow and further talks in Abu Dhabi. Trump said only a few points remain and there is ``no deadline,'' though he called the deal ``very close.'' Macron and EU leaders urged continued pressure and said Russia shows ``no willingness'' for a ceasefire.

\item \textbf{Russia says talks to end Ukraine war `serious' but rules out concessions}
(\emph{Al Jazeera}, November 26, 2025; relevance: 3.0).

Russia says U.S.-brokered talks to end the Ukraine war are ``serious'' and ongoing, but it rules out major concessions. Kremlin spokesman Dmitry Peskov says the process is underway while Russia would not surrender its key positions. The U.S. has pushed a leaked 28-point plan and plans to send top negotiator Steve Witkoff to Moscow. Ukraine and EU allies warned the U.S. proposal reflects Russian demands, including territory, military limits, and NATO barriers. EU foreign policy chief Kaja Kallas said there is ``zero indication'' Russia is ready for a ceasefire and called for additional sanctions.

\item \textbf{Putin accuses Europeans of sabotaging U.S.-led peace efforts in Ukraine}
(\emph{PBS}, December 2, 2025; relevance: 3.0).

Putin accused Kyiv's European allies of sabotaging U.S.-led efforts to end the war, saying they ``don't have a peace agenda'' and are ``on the side of the war.'' He claimed Europeans amend proposals with demands unacceptable to Russia, blocking progress and then blaming Moscow. Putin also said Russia is ready to respond if Europe starts a war. Negotiations involve U.S. envoy Steve Witkoff and Jared Kushner meeting Putin, while Zelenskyy waits for reports from the U.S. talks in Moscow; he said steps depend on signals and results are needed.

\end{enumerate}

\subsection{Example 2: India--Russia Payment Mechanism}

\noindent\textbf{Question.}
Will a joint official statement from the 23rd India--Russia Annual Summit confirm the establishment of a new rupee--ruble alternative payment mechanism to bypass the SWIFT system by January 15, 2026?

\medskip
\noindent\textbf{Prediction date.}
December 4, 2025.

\medskip
\noindent\textbf{Resolution criteria.}
The question resolves to \emph{Yes} if the official joint statement or a specific bilateral agreement issued during or immediately following the 23rd India--Russia Annual Summit, held in or around December 2025, explicitly announces the implementation of a national-currency-based alternative payment system designed to facilitate bilateral trade outside of the SWIFT network. The announcement must confirm that the system is ready for use or being actively deployed, not merely ``under discussion.'' Verification is based on official government releases from the Indian Ministry of External Affairs or the Kremlin.

\medskip
\noindent\textbf{Close date.}
January 15, 2026.

\medskip
\noindent\textbf{Context.}
Recent news articles relevant to the question were summarized as follows:

\begin{enumerate}
\small

\item \textbf{What Putin-Modi Meeting Signals: Defence Deals, Oil Diplomacy And A Test Of India's Strategic Balancing}
(\emph{ETV Bharat}, December 3, 2025; relevance: 4.0).

Putin's December 4--5 India visit for the 23rd India--Russia Annual Summit is expected to yield defence, trade and energy outcomes amid U.S. pressure on India's tariffs and Russian crude purchases. The MEA says it will set the vision for the ``Special and Privileged Strategic Partnership.'' Officials expect a joint statement and multiple agreements, including rupee--ruble local-currency settlement, scaling of mechanisms using Special Rupee Vostro Accounts, and oil discounts, plus nuclear and energy cooperation. The article does not mention a SWIFT bypass or a new, deploy-ready alternative payment mechanism beyond existing rupee--ruble pilots.

\item \textbf{Putin's Delhi Gambit: How a High-Stakes Visit Could Redraw India's Geopolitical Map}
(\emph{South Asian Herald}, November 30, 2025; relevance: 4.0).

Putin's December 4 New Delhi visit at the 23rd India--Russia Summit will test India's ``sovereignty-first'' policy amid its lowest U.S. ties in two decades. Oil is central: since mid-2022, Russian crude drives India---about 35\% of imports and rising to 1.6--1.8 mbpd in FY24--25---after \$10--\$14/bbl discounts. November 21, 2025 sanctions have cut December Russian crude to less than 400,000 bpd, prompting Putin to seek new rupee--ruble clearing channels routed through non-sanctioned banks, plus long-term energy and defense negotiations. U.S. tariffs and CAATSA threats loom.

\item \textbf{Russia and India To Discuss Linking National Payment Systems, Putin To Visit Delhi December 4/5}
(\emph{Russia's Pivot to Asia}, November 29, 2025; relevance: 3.0).

Putin's December 4--5 state visit to India is expected to feature plans to link Russia's and India's payment systems to bypass SWIFT. Deputy Foreign Minister Andrey Rudenko said discussions with India's Jaishankar involved connecting Russia's SBP with India's UPI to enable Mir--RuPay transactions, with hopes for mutual recognition to ease card and payment issues after Western sanctions. Alexey Kupriyanov noted pairing Russia's Faster Payments System and UPI to allow mutual settlements via Mir/RuPay, potentially via QR codes or e-wallets and reducing intermediary fees. The article cites increased trade using rubles and rupees but provides no timeline for implementation by January 15, 2026.

\item \textbf{De-dollarization tipping point as multipolar finance takes hold}
(\emph{Asia Times}, November 10, 2025; relevance: 2.0).

On November 4, Russia's Finance Minister Anton Siluanov said 99.1\% of Russia--China trade is now settled in rubles and yuan outside the Western financial system. The article argues this reflects a shift away from dollar dominance driven by Western sanctions after the Ukraine war, which blocked Russian dollar settlements and forced Russia to rewire trade payment channels. It says cross-border payments can increasingly rely on local institutions and systems such as China's CIPS, bypassing SWIFT. It also claims India-related trade is similarly shifting to local currencies and mentions BRICS aims for a SWIFT-bypassing payments system.

\item \textbf{A reality check for BRICS and the lofty dedollarisation agenda}
(\emph{Lowy Institute}, November 18, 2025; relevance: 2.0).

The article says BRICS members are increasingly using local-currency settlements in bilateral trade to reduce reliance on the U.S. dollar, but a unified BRICS challenge to dollar hegemony remains distant. It notes Iran seeks crypto to bypass sanctions; Russia and China report settling 99.1\% of trade in rubles and yuan and stress banking and payment cooperation; China and Brazil agreed to remove the dollar as an intermediary. India, however, has no plans for a BRICS currency, stressing stability and warning against a currency shared with China.

\end{enumerate}

\section{Bootstrap Confidence Intervals}
\label{app:bootstrap-confidence-intervals}

Table~\ref{tab:bootstrap-confidence-intervals} reports paired, question-level
bootstrap differences in Brier score relative to the base model under the
primary pooled five-rollout evaluation. Each bootstrap resample draws held-out
questions with replacement and retains all five sampled forecasts associated
with each selected question. Differences are computed as the reward-trained variant
minus the base model. Negative values therefore indicate lower Brier score.
Confidence intervals are shown at the 1\%, 5\%, and 10\% significance levels.

\begin{table}[htbp]
\centering
\caption{Paired bootstrap differences in Brier score relative to the base model.}
\label{tab:bootstrap-confidence-intervals}
\begin{tabular}{lrrrrrrr}
\toprule
Comparison & Difference & \multicolumn{2}{c}{99\% CI} & \multicolumn{2}{c}{95\% CI} & \multicolumn{2}{c}{90\% CI} \\
\cmidrule(lr){3-4} \cmidrule(lr){5-6} \cmidrule(lr){7-8}
& & Lower & Upper & Lower & Upper & Lower & Upper \\
\midrule
Log -- Base & -0.0207 & -0.0341 & -0.0082 & -0.0304 & -0.0116 & -0.0289 & -0.0131 \\
Brier -- Base & -0.0213 & -0.0346 & -0.0085 & -0.0314 & -0.0118 & -0.0298 & -0.0132 \\
Spherical -- Base & -0.0181 & -0.0322 & -0.0046 & -0.0287 & -0.0079 & -0.0271 & -0.0096 \\
Beta$(8,8)$ -- Base & -0.0130 & -0.0290 & 0.0027 & -0.0258 & -0.0013 & -0.0236 & -0.0033 \\
Beta$(2,8)$ -- Base & -0.0184 & -0.0308 & -0.0065 & -0.0275 & -0.0096 & -0.0261 & -0.0110 \\
\bottomrule
\end{tabular}
\end{table}

\section{Detailed Results by Evaluation Method}
\label{app:detailed-results}

This appendix reports detailed forecasting and bias--information--noise (BIN)
results under three evaluation specifications: single-sample inference
($n=1$), median-of-five aggregation, and the primary pooled five-rollout
evaluation. For median-of-five aggregation, the final forecast for each question
is the median of five independently sampled probabilities. For the pooled
five-rollout evaluation, all five sampled forecast--outcome pairs are retained
when computing forecast-level metrics, while AUC-ROC is computed separately for
each rollout set and averaged across the five sets. The single-sample and
median-of-five evaluations contain one forecast per question ($N=965$), whereas
the pooled five-rollout evaluation contains all five forecasts per question
($N=4{,}825$). These analyses assess the sensitivity of the main conclusions to
the treatment of stochastic model outputs.

\subsection{Sensitivity to Evaluation Method}

Tables~\ref{tab:n1-vs-ensemble-n1}--\ref{tab:n1-vs-ensemble-pooled_5}
compare forecasting metrics under single-sample inference, median-of-five
aggregation, and pooled five-rollout evaluation. The main performance patterns
are broadly similar across evaluation methods, although median-of-five
aggregation generally improves forecasting performance and changes some
relative rankings among reward variants.

\begin{table}[t!]
\centering
\caption{Metrics for $n{=}1$ forecasts.}
\label{tab:n1-vs-ensemble-n1}
\begin{tabular}{lrrrrrrrrr}
\toprule
Model & Brier & ECE & Log score & AUC & Spherical & Beta(2,8) & Beta(8,8) & Avg tokens & Missing \% \\
\midrule
Base & 0.1867 & 0.1101 & -0.5583 & 0.7292 & 0.7893 & -0.1063 & -0.1314 & 644.0 & 0.10 \\
Log & 0.1665 & 0.0421 & -0.5170 & 0.7431 & 0.8151 & -0.0991 & -0.1105 & 457.4 & 0.00 \\
Brier & 0.1655 & 0.0650 & -0.5160 & 0.7557 & 0.8164 & -0.1000 & -0.1093 & 1087.2 & 0.00 \\
Spherical & 0.1654 & 0.0545 & -0.5193 & 0.7437 & 0.8170 & -0.1000 & -0.1085 & 370.7 & 0.00 \\
Beta(8,8) & 0.1732 & 0.0951 & -0.5606 & 0.7261 & 0.8102 & -0.1111 & -0.1107 & 2671.4 & 0.31 \\
Beta(2,8) & 0.1653 & 0.0396 & -0.5149 & 0.7444 & 0.8167 & -0.0976 & -0.1093 & 779.7 & 0.00 \\
\bottomrule
\end{tabular}
\end{table}

\begin{table}[t!]
\centering
\caption{Metrics for median-of-5 ensemble forecasts.}
\label{tab:n1-vs-ensemble-median_of_5}
\begin{tabular}{lrrrrrrrrr}
\toprule
Model & Brier & ECE & Log score & AUC & Spherical & Beta(2,8) & Beta(8,8) & Avg tokens & Missing \% \\
\midrule
Base & 0.1788 & 0.1071 & -0.5408 & 0.7415 & 0.7994 & -0.1045 & -0.1224 & 662.6 & 0.00 \\
Log & 0.1627 & 0.0465 & -0.5050 & 0.7521 & 0.8195 & -0.0974 & -0.1072 & 454.3 & 0.00 \\
Brier & 0.1610 & 0.0602 & -0.5029 & 0.7707 & 0.8214 & -0.0958 & -0.1063 & 1089.4 & 0.00 \\
Spherical & 0.1648 & 0.0594 & -0.5180 & 0.7467 & 0.8178 & -0.0991 & -0.1077 & 362.8 & 0.00 \\
Beta(8,8) & 0.1701 & 0.0957 & -0.5382 & 0.7487 & 0.8128 & -0.1069 & -0.1097 & 2691.9 & 0.00 \\
Beta(2,8) & 0.1645 & 0.0458 & -0.5121 & 0.7486 & 0.8176 & -0.0979 & -0.1087 & 790.4 & 0.00 \\
\bottomrule
\end{tabular}
\end{table}

\begin{table}[t!]
\centering
\caption{Metrics for pooled-5 forecasts (AUC averaged across rollouts).}
\label{tab:n1-vs-ensemble-pooled_5}
\begin{tabular}{lrrrrrrrrr}
\toprule
Model & Brier & ECE & Log score & AUC & Spherical & Beta(2,8) & Beta(8,8) & Avg tokens & Missing \% \\
\midrule
Base & 0.1861 & 0.1099 & -0.5585 & 0.7248 & 0.7903 & -0.1067 & -0.1297 & 662.6 & 0.02 \\
Log & 0.1653 & 0.0434 & -0.5132 & 0.7407 & 0.8165 & -0.0993 & -0.1093 & 454.3 & 0.00 \\
Brier & 0.1648 & 0.0568 & -0.5185 & 0.7511 & 0.8177 & -0.0996 & -0.1078 & 1089.4 & 0.00 \\
Spherical & 0.1680 & 0.0538 & -0.5273 & 0.7353 & 0.8144 & -0.1017 & -0.1094 & 362.8 & 0.02 \\
Beta(8,8) & 0.1731 & 0.0954 & -0.5536 & 0.7366 & 0.8098 & -0.1093 & -0.1113 & 2691.9 & 0.27 \\
Beta(2,8) & 0.1677 & 0.0449 & -0.5223 & 0.7372 & 0.8141 & -0.1003 & -0.1107 & 790.4 & 0.00 \\
\bottomrule
\end{tabular}
\end{table}

\subsection{BIN Contributions Across Evaluation Methods}

Table~\ref{tab:n1-vs-ensemble-compare-bin-contributions} compares the estimated
bias, information, and noise contributions under single-sample inference,
median-of-five aggregation, and pooled five-rollout evaluation. Several patterns
are stable across evaluation methods: the Brier variant has the largest
information contribution, while the log variant consistently shows a positive
noise contribution. Notable changes occur for Beta$(2,8)$ and Beta$(8,8)$:
Beta$(2,8)$ shifts from a small positive noise contribution under single-sample
evaluation to a negative contribution under median-of-five evaluation and back
to a small positive contribution under pooled evaluation, while Beta$(8,8)$
shows substantially smaller bias contributions under median-of-five and pooled
evaluation than under $n=1$.

\begin{table}[t!]
\centering
\caption{BIN contributions by evaluation method.}
\label{tab:n1-vs-ensemble-compare-bin-contributions}
\begin{tabular}{lrrrrrrrrr}
\toprule
& \multicolumn{3}{c}{Bias (\%)} 
& \multicolumn{3}{c}{Noise (\%)} 
& \multicolumn{3}{c}{Information (\%)} \\
\cmidrule(lr){2-4} 
\cmidrule(lr){5-7} 
\cmidrule(lr){8-10}
Reward 
& $n{=}1$ & Median of 5 & Pooled 5
& $n{=}1$ & Median of 5 & Pooled 5
& $n{=}1$ & Median of 5 & Pooled 5 \\
\midrule
Log
& 6.55 & 5.56 & 6.06
& 2.15 & 1.32 & 2.60
& 0.83 & 1.16 & 1.29 \\

Brier
& 5.50 & 4.51 & 5.75
& 0.65 & 0.39 & 0.44
& 4.11 & 3.88 & 3.70 \\

Spherical
& 6.18 & 5.56 & 6.35
& 1.09 & -1.52 & -0.38
& 2.05 & 2.02 & 2.16 \\

Beta$(2,8)$
& 6.87 & 6.05 & 6.63
& 0.10 & -0.66 & 0.36
& 3.06 & 1.83 & 1.90 \\

Beta$(8,8)$
& 4.50 & 1.89 & 3.73
& -2.69 & -1.50 & -1.57
& 3.48 & 3.50 & 3.68 \\
\bottomrule
\end{tabular}
\end{table}

\section{Training-Seed Robustness}
\label{app:seed-robustness}

To assess sensitivity to training stochasticity, we repeated the Log and
Brier reward conditions across three training seeds, holding all other
training and evaluation settings fixed.
Table~\ref{tab:multi-seed-forecasting-results} reports the resulting
forecasting metrics. Log has slightly better mean Brier score, log score,
and ECE, while mean AUC-ROC is identical across the two conditions.

\begin{table}[t!]
\centering
\caption{Forecasting performance across training seeds for the Log and Brier reward conditions.}
\label{tab:multi-seed-forecasting-results}
\begin{tabular}{lrrrr}
\toprule
Model & Brier $\downarrow$ & ECE $\downarrow$ & Log score $\uparrow$ & AUC-ROC $\uparrow$ \\
\midrule
Log, seed 1   & 0.1638 & 0.0336 & -0.5096 & 0.7431 \\
Log, seed 2   & 0.1644 & 0.0312 & -0.5099 & 0.7401 \\
Log, seed 3   & 0.1670 & 0.0632 & -0.5184 & 0.7354 \\
\textbf{Log, mean} & \textbf{0.1651} & \textbf{0.0426} & \textbf{-0.5126} & \textbf{0.7395} \\
Brier, seed 1 & 0.1682 & 0.0732 & -0.5263 & 0.7331 \\
Brier, seed 2 & 0.1669 & 0.0589 & -0.5268 & 0.7334 \\
Brier, seed 3 & 0.1635 & 0.0645 & -0.5195 & 0.7521 \\
\textbf{Brier, mean} & \textbf{0.1662} & \textbf{0.0655} & \textbf{-0.5242} & \textbf{0.7395} \\
\bottomrule
\end{tabular}
\end{table}

The BIN decomposition shows a clearer distinction. Using the three seeds
as separate experts and the probit mean of five rollouts, Brier has a
larger information contribution than Log, while Log has more favorable
bias and noise contributions.

\begin{table}[t!]
\centering
\caption{BIN contributions for Brier relative to Log using three training seeds as experts.}
\label{tab:multi-seed-bin-components}
\begin{tabular}{lrrr}
\toprule
Treatment vs.\ control & Bias (\%) & Noise (\%) & Information (\%) \\
\midrule
Brier vs.\ Log & -1.19 & -2.28 & 2.46 \\
\bottomrule
\end{tabular}
\end{table}

\begin{table}[t!]
\centering
\caption{Posterior probabilities that Brier improves on Log in each BIN component.}
\label{tab:multi-seed-bin-vs-control-probs}
\begin{tabular}{lrrr}
\toprule
Treatment vs.\ control &
$P(\text{less }|\mathrm{bias}|)$ &
$P(\text{more information})$ &
$P(\text{less noise})$ \\
\midrule
Brier vs.\ Log & $<0.001$ & $>0.999$ & $<0.001$ \\
\bottomrule
\end{tabular}
\end{table}

These results suggest that while small aggregate-performance differences
are seed-sensitive, the qualitative BIN distinction between Log and Brier
is robust across the repeated runs.

\end{document}